\documentclass{article}
\usepackage{spconf,amsmath,graphicx}
\usepackage{helvet} % DO NOT CHANGE THIS
\usepackage{courier}  % DO NOT CHANGE THIS
\usepackage{graphicx} % DO NOT CHANGE THIS
\usepackage{caption} % DO NOT CHANGE THIS AND DO NOT ADD ANY OPTIONS TO IT

\usepackage{graphicx}
\usepackage{booktabs}
\usepackage{subfigure}
\usepackage{algorithm}
\usepackage{algorithmic}
\usepackage{color}

\usepackage{microtype}
\usepackage{amsmath, amssymb, amscd, amsfonts}
\usepackage{IEEEtrantools}
\usepackage{soul}
\usepackage{url}
\usepackage{amsmath}
\usepackage{algorithm}
\usepackage{algorithmic}
\usepackage{amssymb}
\usepackage{amsfonts} 
\usepackage{multirow}
\usepackage{boldline}
\usepackage{hhline}
\usepackage{xcolor}

\usepackage{makecell}
\usepackage{mathrsfs}
\usepackage[normalem]{ulem}
\useunder{\uline}{\ul}{}
\usepackage[inline]{enumitem}
\usepackage{enumitem}
\usepackage{bbding}
\usepackage{pifont}

\usepackage{indentfirst}
\usepackage[T1]{fontenc}
\usepackage[utf8]{inputenc}

\usepackage{microtype}
\title{Tell Model Where to Attend: Improving Interpretability of Aspect-Based Sentiment Classification via Small Explanation Annotations}
\name{Zhenxiao Cheng$^\dagger$, Jie Zhou$^\dagger$\sthanks{Jie Zhou is the corresponding author, jzhou@ica.stc.sh.cn.}, Wen Wu$^\dagger$, Qin Chen$^\dagger$, Liang He$^\dagger$}
\address{$^\dagger$School of Computer Science and Technology, East China Normal University, Shanghai, China 
}
\begin{document}
%\ninept
%
\maketitle
\begin{abstract}

Gradient-based explanation methods play an important role in the field of interpreting complex deep neural networks for NLP models. However, the existing work has shown that the gradients of a model are unstable and easily manipulable, which impacts the model's reliability largely. %, which limits the practicability largely.
According to our preliminary analyses, we also find the interpretability of gradient-based methods is limited for complex tasks, such as aspect-based sentiment classification (ABSC).
In this paper, we propose an \textbf{I}nterpretation-\textbf{E}nhanced \textbf{G}radient-based framework for \textbf{A}BSC via a small number of explanation annotations, namely \texttt{{IEGA}}. 
Particularly, we first calculate the word-level saliency map based on gradients to measure the importance of the words in the sentence towards the given aspect. Then, we design a gradient correction module to enhance the model's attention on the correct parts (e.g., opinion words). Our model is model agnostic and task agnostic so that it can be integrated into the existing ABSC methods or other tasks. Comprehensive experimental results on four benchmark datasets show that our \texttt{IEGA} can improve not only the interpretability of the model but also the performance and robustness.
\end{abstract}

\section{Introduction}
\label{sec:intro}
Interpreting complex deep neural networks to understand the reasoning behind the decision of NLP models has attached much attention \cite{danilevsky-etal-2020-survey,zhou2021attending}. 
% Interpreting complex machine learning models, such as deep neural networks, remains a challenge. 
Understanding how such models work is an important research with wide applications such as deployment \cite{ribeiro2016should} and helping developers improve the quality of the models \cite{wallace2019allennlp}. 
Recently, post-hoc explanation techniques have been widely used, categorized into black-box and white-box methods. 
In this paper, we improve upon the gradient-based explanation method \cite{DBLP:journals/corr/SimonyanVZ13,sundararajan2017axiomatic}, which is one of the main methods in white-box models.
% Our work extends and improves upon the gradient-based method \cite{DBLP:journals/corr/SimonyanVZ13}, a popular technique applicable to many different types of models. Accordingly, we focus on gradient-based methods.
% Post-hoc explanation techniques are helpful for such insights, for example, to evaluate whether a model is doing the ``right thing” before deployment \cite{ribeiro2016should,lundberg2017unified} to increase human trust into black box systems \cite{doshi2017towards}, and to help diagnose model biases \cite{wallace2019allennlp}.
% Gradient-based explanation algorithms (e.g., saliency map visualization) have been widely used in interpreting neural models for many computer version (CV) and natural language processing (NLP) tasks due to their faithfulness. However, there is considerable ambiguity in how to convert dimension-level signed gradient values to word-level weights for NLP tasks. 
Gradient-based explanation method calculates the word importance explainability by estimating the contribution of input sentence $x$ towards output $y$ \cite{aubakirova2016interpreting,karlekar2018detecting,zhou2021hot}. 
It calculates the first derivative of $y$ with respect to $x$ to obtain the saliency map, which is a popular technique applicable to various deep learning models.
% Recent work has also proposed improvements to first-derivative saliency \cite{sundararajan2017axiomatic}. As suggested by its
% name and definition, first-derivative saliency can be
% used to enable feature importance explainability, especially on word/token-level features \cite{aubakirova2016interpreting,karlekar2018detecting}.

\begin{figure}[t!]
% \vspace{-2mm}
\centering
\includegraphics[scale=0.35]{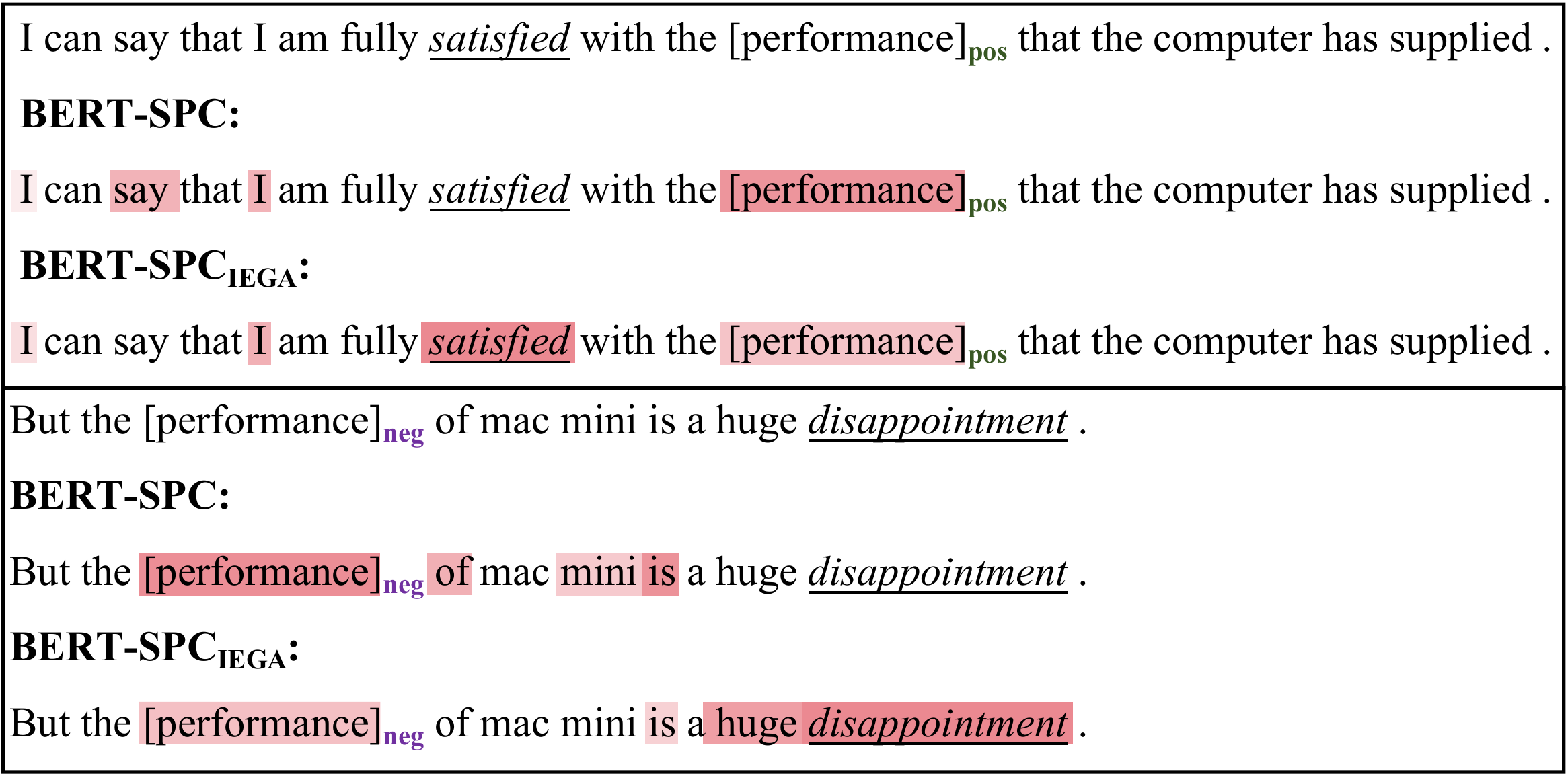}
\vspace{-2mm}
\caption{Two examples of interpretability. The words marked with [] and \underline{underline} are aspects and opinion words respectively. Pos and neg mean positive and negative sentiments of the aspects. Color depth indicates the importance of the words. 
% Interpretation
}
\label{fig:example}
\vspace{-5mm}
\end{figure}

However, the existing literature has shown that the gradient-based model is easily manipulable \cite{wang2020gradient} and unreliable \cite{kindermans2019reliability}. Moreover, we also find that the gradient-based methods perform poorly on complex NLP tasks, such as aspect-based sentiment classification (ABSC). ABSC aims to judge the sentiment polarity of the given aspect in the sentence, which may contain multiple aspects whose sentiments may be opposite. For example, in the sentence "I can say that I am fully satisfied with the performance that the computer has supplied.", the user expresses a positive sentiment towards the aspect "performance" using the opinion words "satisfied" (Fig. \ref{fig:example}). We can find that the BERT-SPC model focuses on the unrelated words (e.g., "say", "I", "of", "is") via the standard gradient. 
Here, we hope the model can capture the most relevant words (e.g., "satisfied", "disappointment") for predicting based on small additional explanation annotations because labeling the fine-grained opinions is expensive.  

\begin{figure*}[t!]
\vspace{-9mm}
\centering
\includegraphics[scale=0.5]{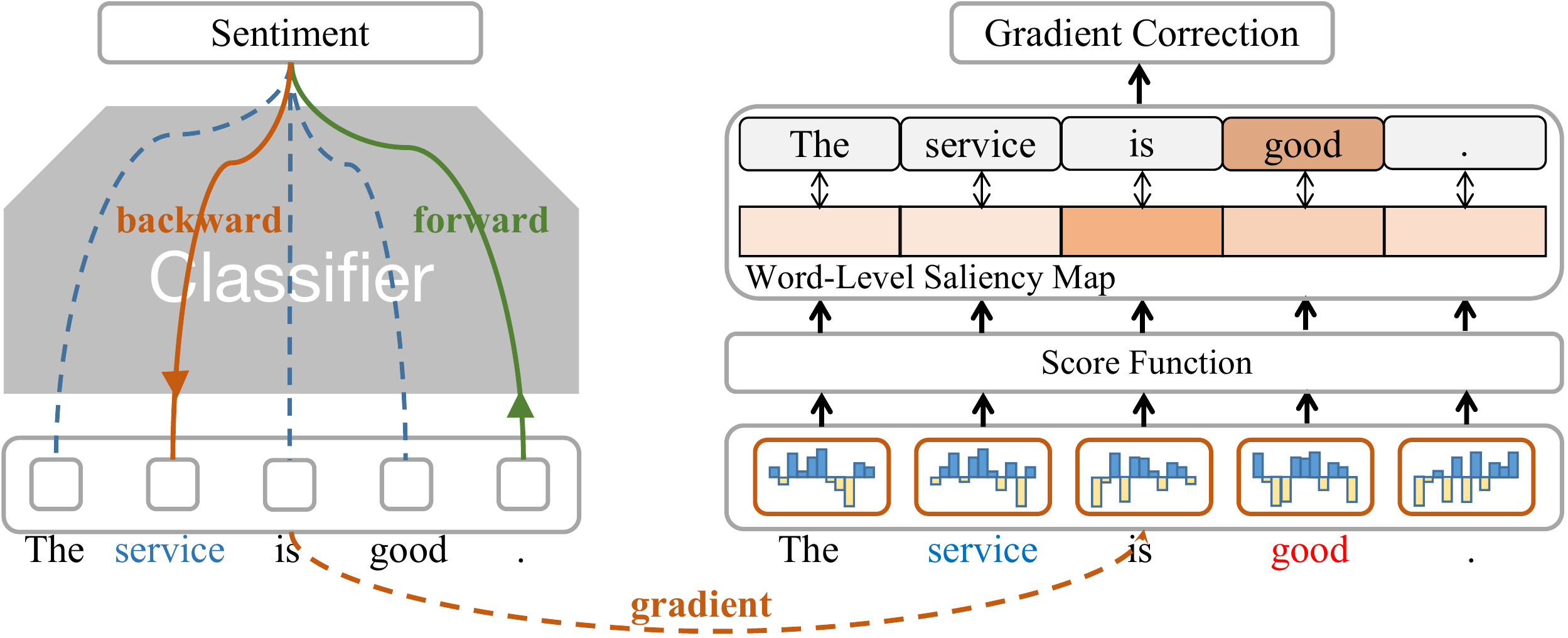}
\vspace{-3mm}
\caption{The architecture of our \texttt{IEGA} proposed model. 
}
\label{fig:framework}
\vspace{-5mm}
\end{figure*}

Particularly, to enhance the model doing the "right thing”, we propose an \textbf{I}nterpretation-\textbf{E}nhanced \textbf{G}randient-based method (\texttt{IEGA}) for \textbf{A}BSC via small annotations. 
First, we use gradients to obtain the word-level saliency map, which measures the importance of the words for the predictions. 
Since labeling the opinion words toward the given aspects is time-consuming, we then aim to guide the model via small explainable annotations. Specifically, we design a gradient correction module to enforce the model to focus on the most related words (e.g., opinion words). 
We also conduct a series of experiments on four popular datasets for the ABSC task. The results indicate that our \texttt{IEGA} can improve not only the interpretability but also the performance of the model. Additionally, we also investigate the robustness of our model. 

The main contributions of this paper can be summarized as follows. 1) We propose a novel framework to improve the interpretation of gradient-based methods for ABSC via a small number of additional explanation samples; 2) We design a gradient correction module to enable the model to capture the most related words based on the word-level saliency map; 3) We conduct extensive experiments to verify the great advantages of our \texttt{IEGA} framework, which can improve the explanation, performance, and robustness of the model.
% \begin{itemize}
%     \item We propose a simple framework to improve the interpretation of gradient-based methods for ABSC via few labelled samples. Specifically, we design a gradient correction module to enable model to capture the most related words.
%     \item We conduct rich experiments to verify the great advantages of our \texttt{IEGA} framework, which can improve the explanation, performance and robustness of the model.
% \end{itemize}

\section{Approaches}
In this paper, we propose an \texttt{IEGA} framework for ABSC to enhance the interpretation of the model using small annotations (Fig. \ref{fig:framework}). First, we compute the gradients of $y$ with respect to $x$ based on a sentiment classifier. Second, we introduce a gradient correction module to make the word-level saliency map obtained by gradients close to the truth distributions of opinion words.

Formally, given a sentence $s=\{w_1, ..., w_i, ..., w_{|s|}\}$ and an aspect $a$, where aspect $a$ is a subsequence of $s$, $w_i$ is the $i$-th word in $s$, and $|s|$ is the number of the words in the sentence. The word embeddings of the sentence $s$ are $x=\{x_1, ..., x_i, ..., x_{|s|}\}$, where $x_i$ is the word embedding of $w_i$. This task aims to predict the sentiment polarity $y \in \{P, N, O\}$ of the sentence towards the given aspect $a$. $P$, $N$, $O$ represent positive, negative, and neutral, respectively. Moreover, we aim to explain the model by extracting the opinion words $o$ that express the sentiment w.r.t the aspect $a$.

\subsection{Gradient Calculation}
First, we train a sentiment classifier for ABSC, which aims to infer the sentiment of the sentence w.r.t. the given aspect.
Let $\mathcal{F}$ be a sentiment classifier which predict the sentiment distribution $P(y|s,a)$ based on the sentence $s$ and aspect $a$.
\begin{equation}
    P(y|s,a) = \mathcal{F}(s, a)
\end{equation}

The loss function is the cross-entropy between the predicted probability and the true label,
\begin{equation}
\mathcal{L}_c(y, s, a) = -{\log P(y|s, a)} 
\label{loss}.
\end{equation}

Particularly, the sentiment classifier $\mathcal{F}$ can be any existing aspect-level sentiment classifier models, such as BERT-SPC \cite{xu-etal-2019-bert}, RGAT-BERT \cite{wang-etal-2020-relational}. 

If we slightly perturb the word $w_i$'s embedding $\mathbf{x_i}$ to $\mathbf{x_i}'$ 
with $\|\mathbf{x_i}' - \mathbf{x_i}\|\leq \varepsilon$,
we can use the first-order approximation of $\mathcal{L}_c(y, s, a)_{\mathbf{x_i}}$ to compute the absolute change of the loss function, which indicates the importance of the word $w_i$.
\begin{IEEEeqnarray*}{rl}
    & |\mathcal{L}_c(y, s, a)_{\mathbf{x_i}'} - \mathcal{L}_c(y, s, a)_{\mathbf{x_i}}| \\
    &\approx |\nabla_{\mathbf{x_i}}\mathcal{L}_c(y, s, a)_{\mathbf{x_i}})^T
    (\mathbf{x_i}' - \mathbf{x_i})| \\
    & \leq \|\nabla_{\mathbf{x_i}}\mathcal{L}_c(y, s, a)_{\mathbf{x_i}}\|
\|\mathbf{x_i}' - \mathbf{x_i}\| \\
    & \leq \varepsilon \|\nabla_{\mathbf{x_i}}\mathcal{L}_c(y, s, a)_{\mathbf{x_i}}\|.
\end{IEEEeqnarray*}

Thus, it is a natural approximation to metric what tokens are most relevant for a prediction using the gradient \cite{DBLP:journals/corr/SimonyanVZ13,baehrens2010explain}. 
The gradient of the $i$-th word $w_i$ is calculated as follows:
\begin{equation}
g_i = \nabla_{\mathbf{x_i}}\mathcal{L}_c(y, s, a)_{\mathbf{x_i}} = \frac{\partial \mathcal{F}\left(s, a\right)}{\partial x_i}
\end{equation}
which can be computed automatically via machine learning frameworks, such as Pytorch and Tensorflow.

\subsection{Gradient Correction}
% \subsection{Gradient-base Token Attribution}
In order to enhance the model to focus on the correct parts (e.g., opinion words), we introduce a gradient correction module. 
We first calculate the importance of the words in the sentence based on the gradients to obtain the word-level saliency map. 
The magnitude of the gradient's norm 
$\|g_i\|$ could be a sign of how sensitive the 
sentiment label is to $w_i$:
to get the correct prediction we will prefer not
to perturb those words with large gradient norms.
% Therefore, a large gradient norm may also indicate an
% opinion words of the target.
It suggests that words with large gradient norm are contributing 
most towards the correct sentiment label and
might be the opinion words of the aspect.
Thus, we define the attribution for the word $w_i$ as
\begin{equation*}
    \alpha_i = \frac{\mathrm{score}(w_i)}{\sum_{j}(\mathrm{score}(w_j))},
\end{equation*}
where $\mathrm{score}(w_i)=\left|g_i \cdot x_i \right|$ converts the word gradients into weights by the dot product of the gradient $g_i$ and word embedding $x_i$. Gradients (of the output with respect to the input) are a natural analog of the model coefficients for a deep network. Therefore the product of the gradient and feature values is a proper importance score function \cite{DBLP:journals/corr/SimonyanVZ13,baehrens2010explain}.

% Let $f$ be a classifier which takes as input a sequence of embeddings $x=(x_1, x_2, ..., x_n)$. The gradient with respect to the input is often used in analysis methods, which we represent as the normalized gradient attribution vector $a = (a_1, a_2, ..., a_n)$ over the tokens. Similar to past work \cite{feng2018pathologies}, we define the attribution at position $i$ as
% \begin{equation}
%     a_i = \frac{\mathrm{score}(g_i)}{\sum_j \mathrm{score}(g_j)}
% \end{equation}
% where $\mathrm{score}(x)$ converts the word gradients into weights. Most of the existing methods calculate it using a sum.

We use small labeled samples to make the word-level saliency map close to the distributions of opinion words. 
\begin{equation}
\mathcal{L}_{g} = - \sum_{j=1}^{|s|} y^{o}_{j} \alpha_{j}
\end{equation}
where $y^{o}_{j}=1$ if word $w_j$ is opinion word, else $y^{o}_{j} = 0$.

Finally, we add the classification loss $\mathcal{L}_c$ and gradient correction loss $\mathcal{L}_g$ with a weight $\lambda$, $\mathcal{L} = \mathcal{L}_c + \lambda \mathcal{L}_g$.
% \begin{equation}
%     \mathcal{L} = \mathcal{L}_c + \lambda \mathcal{L}_g
% \end{equation}

\vspace{-4mm}
\section{Experiments}
\label{sec:experiments}

\subsection{Experimental Setups}
\label{ssec:experimental setups}
\textbf{Datasets.} 
To verify the effectiveness of \texttt{IEGA}, we conduct extensive experiments on four typical datasets: Res14, Lap14, Res15, and Res16 \cite{fan-etal-2019-target}. We follow the setting of \cite{fan-etal-2019-target}, which labeled the opinion words for each aspect. 
% Table \ref{table:statistics of the datasets} lists the statistics information of the datasets.

% \begin{table}[t!]
% \vspace{-7mm}
% \centering
% \small
% \caption{The statistics of the datasets.}
% \label{table:statistics of the datasets}
% \vspace{-2mm}
% \begin{tabular}{clcccccccc}
% \hlineB{4}
% &Dataset   & Positive & Negative & Neutral & Total      \\
%                          \hline
% \multirow{2}{*}{Lap14} & Training                 & 854       & 579       & 152       & 1585\\
%                        & Testing                  & 304       & 101       & 62        & 467 \\ \hline
% \multirow{2}{*}{Res14} & Training                 & 1806      & 540       & 209       & 2555\\
%                        & Testing                  & 650       & 141       & 59        & 850\\ \hline
% \multirow{2}{*}{Res15} & Training                 & 824       & 219       & 32        & 1075\\
%                        & Testing                  & 292       & 123       & 19        & 434\\ \hline
% \multirow{2}{*}{Res16} & Training                 & 1103      & 345       & 51        & 1499\\
%                        & Testing                  & 359       & 72        & 25        & 456\\
% \hlineB{4}
% \end{tabular}
% \vspace{-2mm}
% \end{table}

\textbf{Metrics. }
To assess the models' performance, we use two popular metrics, Accuracy and Macro-F1. 
To investigate the faithfulness of explanations, follow \cite{wang2020gradient}, we used Mean Reciprocal Rank (MRR) and Hit Rate (HR) to verify whether opinion words get higher scores in attribution. 
% \textcolor[rgb]{0,0,1}{The higher these two metrics are, the more accurate the model is in terms of attribution.}
We also adopt the area over the perturbation curve (AOPC) \cite{nguyen2018comparing,samek2016evaluating}  and Post-hoc Accuracy (Ph-Acc) \cite{pmlr-v80-chen18j}, which are widely used for explanations \cite{chen2020learning}. 
AOPC calculates the average change of accuracy over test data by deleting top $K$ words via the word-level saliency map. 
For Post-hoc Accuracy, we select the top $K$ words based on their importance weights as input to make a prediction and compare it with the ground truth. 
% \textcolor[rgb]{0,0,1}{If the values of a model's AOPC and Post-hoc Accuracy increase, it proves that the interpretability of the model has been improved.}
We set $K=5$ value in our experiments. 
% Note that a well-trained classifier is used for evaluating the performance.

\textbf{Baselines. }
We select four state-of-the-art baselines for aspect-based sentiment classification to investigate the performance: BERT-SPC \cite{xu-etal-2019-bert}, AEN-BERT \cite{song2019attentional}, LCF-BERT \cite{zeng2019lcf}, RGAT-BERT \cite{wang-etal-2020-relational}. For the limitation of space, please see more details about the baselines on the original papers.
% \textbf{BERT-SPC} \cite{xu-etal-2019-bert} simply connects the aspect words with the original sentence to get the input sequences and then input these sequences into BERT for training. 
% \textbf{AEN-BERT} \cite{song2019attentional} employs attention based encoders for the modeling between context and target. 
% \textbf{LCF-BERT} \cite{zeng2019lcf} uses multi-head self-attention to make the model focus on the local context words. 
% \textbf{RGAT-BERT} \cite{wang-etal-2020-relational} adopts a relation-aware graph attention network to encode the new tree structure for sentiment prediction.

\textbf{Implementation Details. }
While conducting our experiments, we adopt the BERT base as the pre-trained model of our sentiment classifier. Some hyperparameters like batch size, maximum epochs, and learning rate are set to 32, 20, and 2e-5. The weight $\lambda$ of gradient correction loss is fixed at 0.01.

\begin{figure}[t!]
% \vspace{-7mm}
\begin{minipage}[b]{0.49\linewidth}
  \centering
  \centerline{\includegraphics[width=4.0cm]{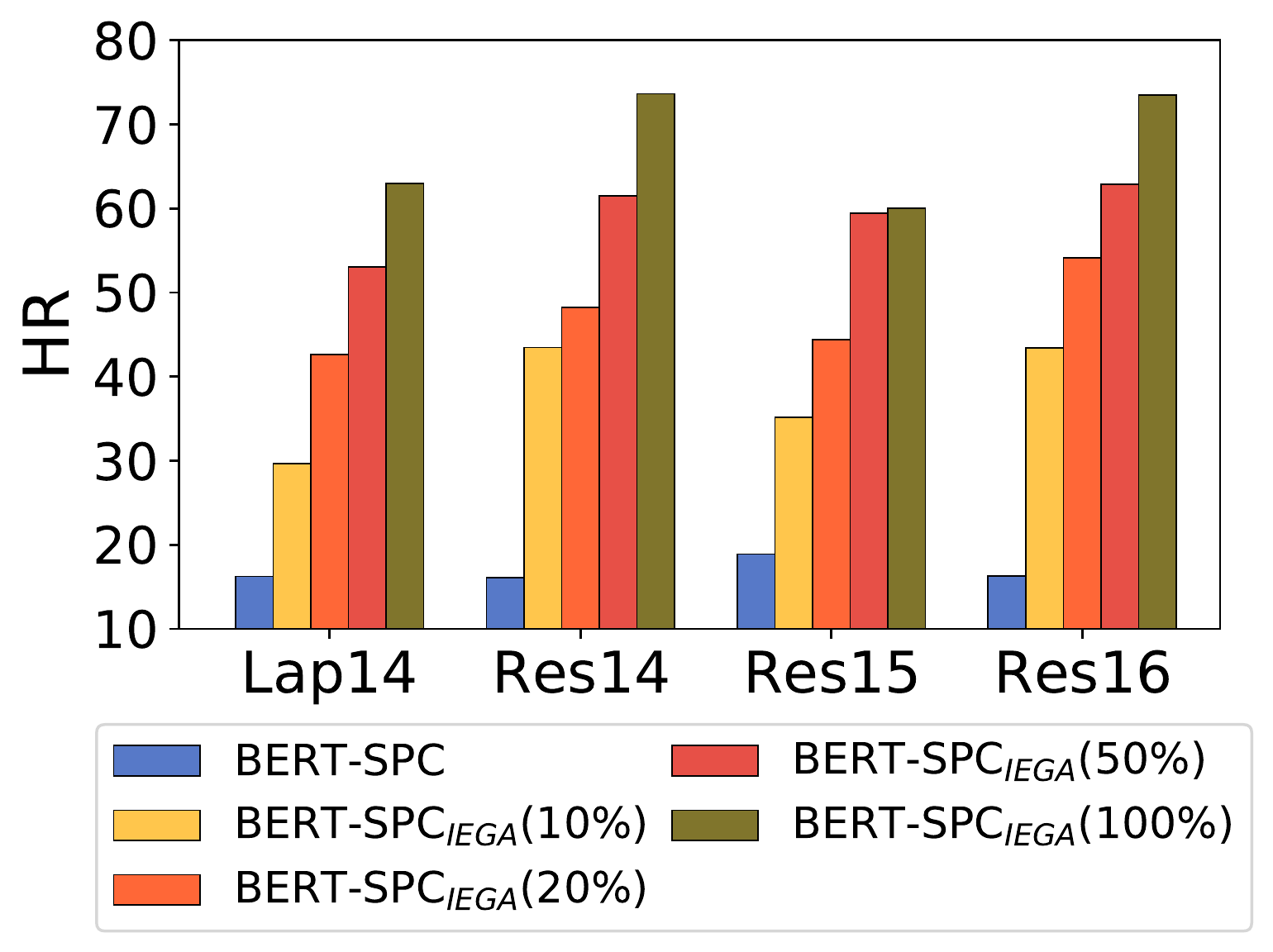}}
%  \vspace{2.0cm}
%   \centerline{(a) Result 1}\medskip
\end{minipage}
\begin{minipage}[b]{0.49\linewidth}
  \centering
  \centerline{\includegraphics[width=4.0cm]{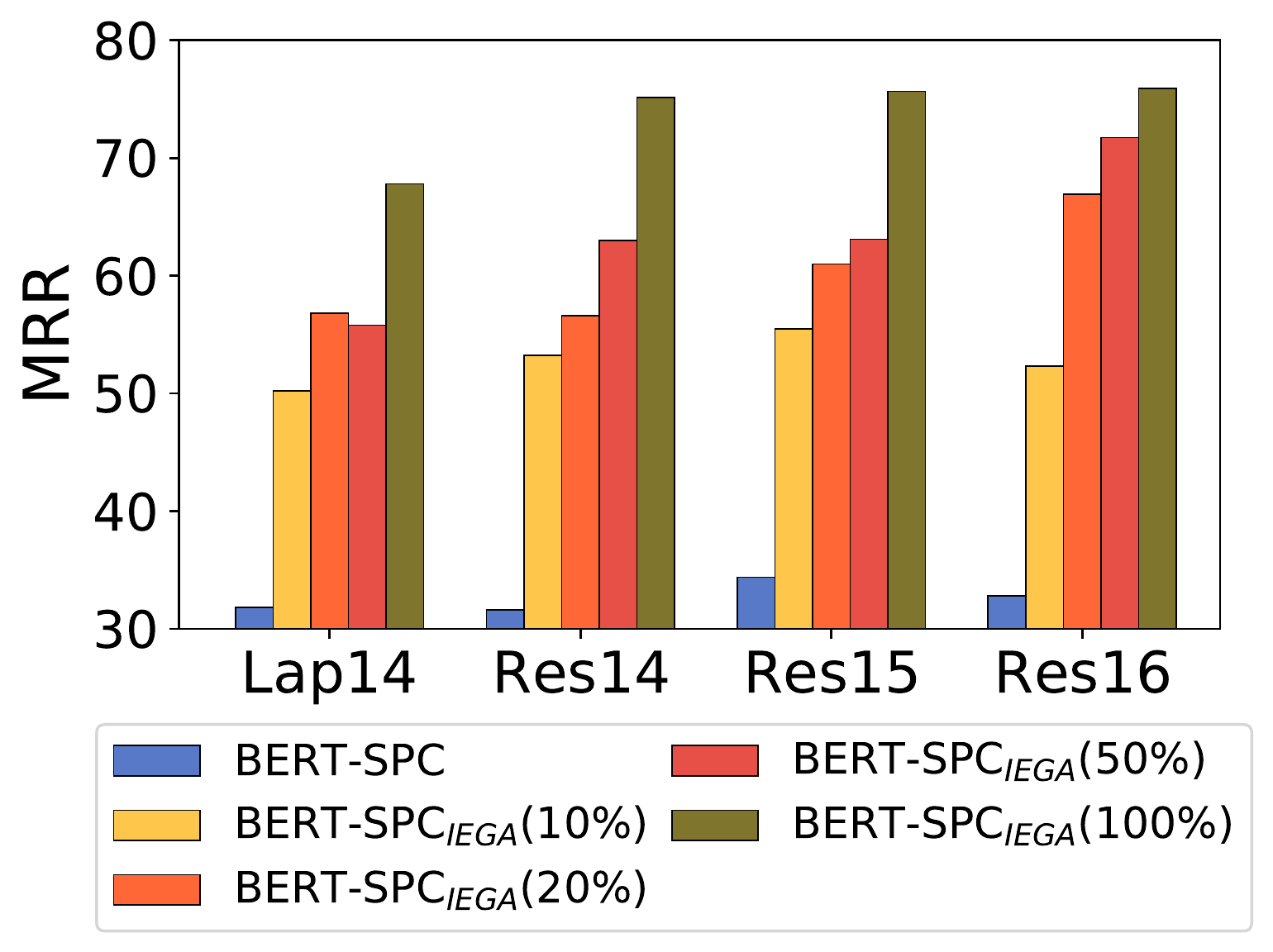}}
%  \vspace{1.5cm}
%   \centerline{(b) MRR}\medskip
\end{minipage}

 %\hfill

\begin{minipage}[b]{0.49\linewidth}
  \centering
  \centerline{\includegraphics[width=4.0cm]{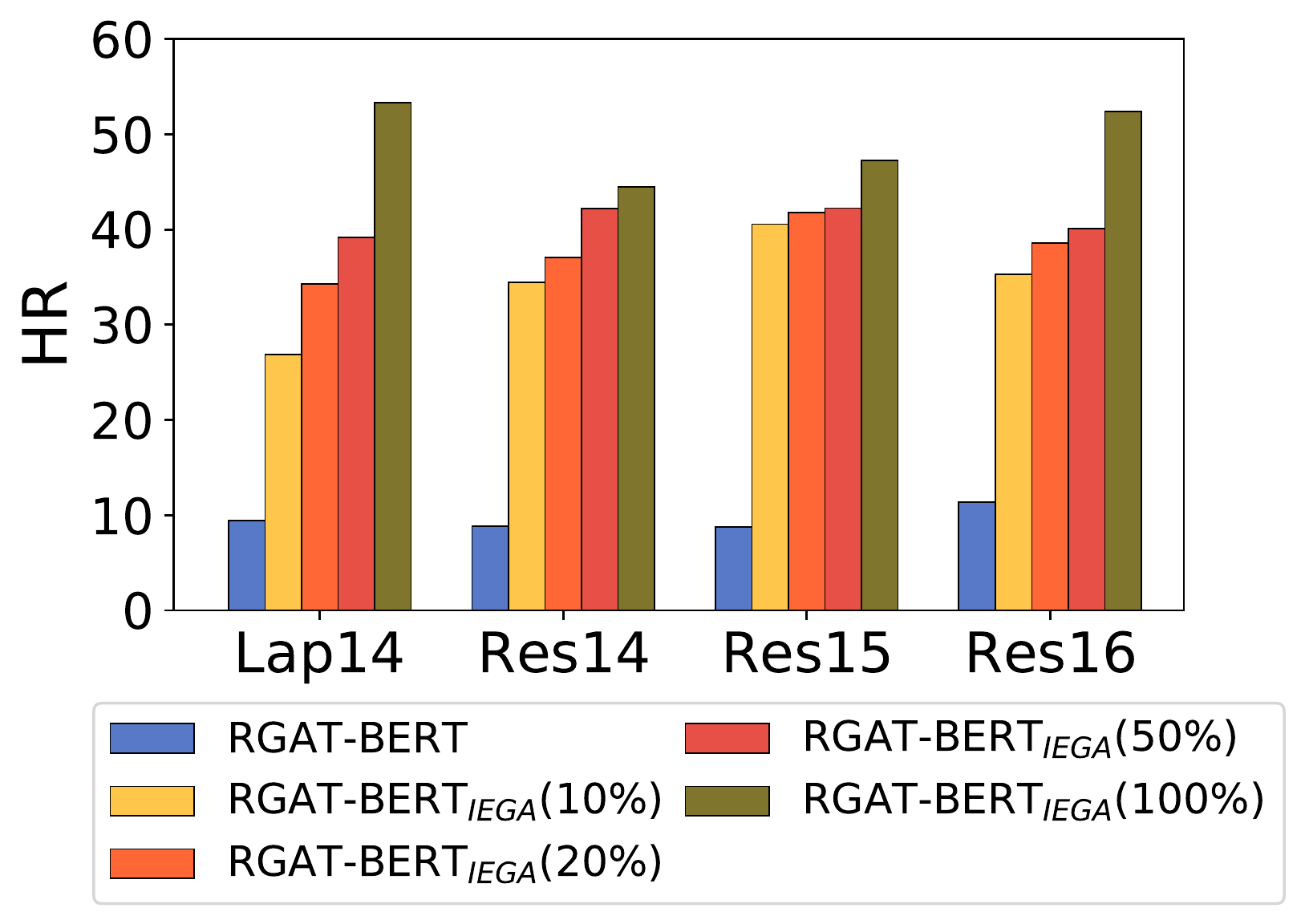}}
%  \vspace{1.5cm}
%   \centerline{(c) Result 4}\medskip
\end{minipage}
\begin{minipage}[b]{0.49\linewidth}
  \centering
  \centerline{\includegraphics[width=4.0cm]{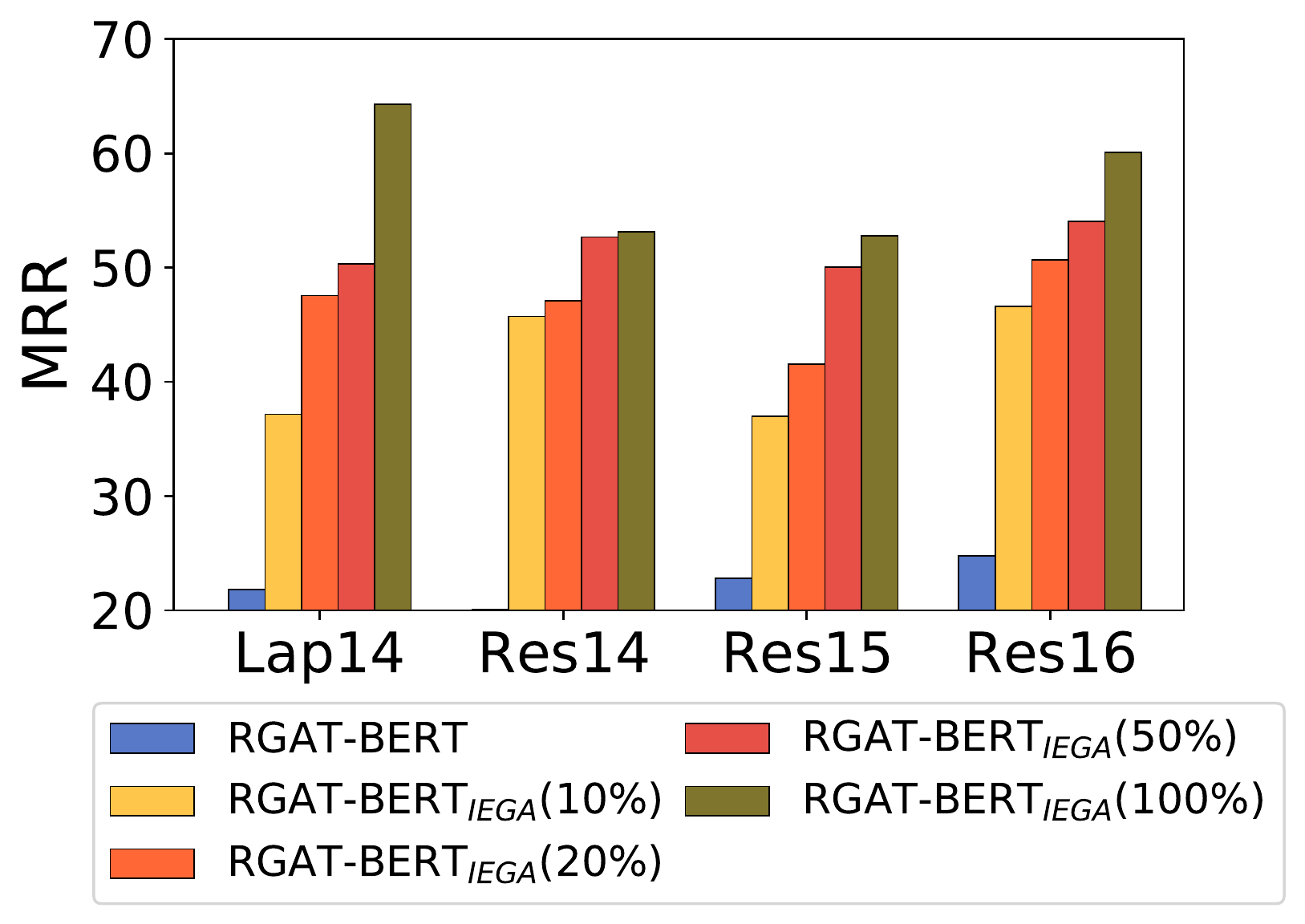}}
%  \vspace{1.5cm}
%   \centerline{(c) Result 4}\medskip
\end{minipage}
\vspace{-1mm}
\caption{The results of interpretability.}
\label{fig:results of interpretability}
\vspace{-5mm}
\end{figure}

%new tabele update on 2022.10.06
\begin{table*}[t!]
\vspace{-9mm}
\centering
\small
\caption{The results of interpretability in terms of AOPC and Post-hoc Acc. 
%The larger the value of AOPC, the better the explanations of the models. 
}
\label{table:results of interpretability_2}
\vspace{-2mm}
\setlength{\tabcolsep}{0.8mm}{\begin{tabular}{lllllllll}
\hlineB{4}
\multirow{2}{*}{Model}         & \multicolumn{2}{c}{Lap14} & \multicolumn{2}{c}{Res14} & \multicolumn{2}{c}{Res15} & \multicolumn{2}{c}{Res16} \\ %\cline{2-9} 
                               & AOPC$\uparrow$    & Ph Acc$\uparrow$     & AOPC$\uparrow$    & Ph Acc$\uparrow$     & AOPC$\uparrow$    & Ph Acc$\uparrow$     & AOPC$\uparrow$    & Ph Acc$\uparrow$   \\ \hline
BERT-SPC                       & 07.24    & 33.70           & 07.95    & 56.87           & 06.50    & 64.84           & 03.82    & 65.44           \\
BERT-SPC$_\texttt{IEGA}$ (10\%)   & 15.04\scriptsize {(+7.80)}    & 42.18\scriptsize {(+8.48)}           & 11.13\scriptsize {(+3.18)}    & 69.18\scriptsize {(+12.31)}           & 08.26\scriptsize {(+1.76)}    & 70.78\scriptsize {(+5.94)}           & 10.48\scriptsize {(+6.66)}    & 75.49\scriptsize {(+10.05)}           \\ 
BERT-SPC$_\texttt{IEGA}$ (20\%)   & 18.54\scriptsize {(+11.30)}    & 42.38\scriptsize {(+8.68)}           & 12.35\scriptsize {(+4.40)}    & 74.17\scriptsize {(+17.30)}           & 09.65\scriptsize {(+3.15)}    & 71.98\scriptsize {(+7.14)}           & 11.49\scriptsize {(+7.67)}    & 76.27\scriptsize {(+10.83)}           \\
BERT-SPC$_\texttt{IEGA}$ (50\%)   & 18.57\scriptsize {(+11.33)}   & 55.67\scriptsize {(+21.97)}           & 16.78\scriptsize {(+8.83)}    & 75.29\scriptsize {(+18.42)}           & 10.50\scriptsize {(+4.00)}    & 73.50\scriptsize {(+8.66)}           & 12.41\scriptsize {(+8.59)}    & 78.17\scriptsize {(+12.73)}           \\
BERT-SPC$_\texttt{IEGA}$ (100\%)  & 21.90\scriptsize {(+14.66)}   & 59.73\scriptsize {(+26.03)}           & 20.47\scriptsize {(+12.52)}   & 76.12\scriptsize {(+19.25)}           & 11.98\scriptsize {(+5.48)}    & 74.65\scriptsize {(+9.81)}          & 15.87\scriptsize {(+12.05)}    & 79.82\scriptsize {(+14.38)}           \\
\hline
RGAT-BERT                      & 05.28    & 54.38           & 12.70   & 51.64           & 08.06    & 64.51           & 08.31    & 71.05           \\
RGAT-BERT$_\texttt{IEGA}$ (10\%)  & 13.58\scriptsize {(+8.30)}    & 66.52\scriptsize {(+12.14)}           & 15.67\scriptsize {(+2.97)}   & 63.29\scriptsize {(+11.65)}           & 16.07\scriptsize {(+8.01)}   & 72.98\scriptsize {(+8.47)}           & 12.71\scriptsize {(+4.40)}    & 80.04\scriptsize {(+8.99)}           \\
RGAT-BERT$_\texttt{IEGA}$ (20\%)  & 13.92\scriptsize {(+8.64)}    & 67.89\scriptsize {(+13.51)}           & 15.88\scriptsize {(+3.18)}   & 68.82\scriptsize {(+17.18)}           & 17.97\scriptsize {(+9.91)}   & 77.89\scriptsize {(+13.38)}           & 12.80\scriptsize {(+4.49)}    & 80.70\scriptsize {(+9.65)}                \\
RGAT-BERT$_\texttt{IEGA}$ (50\%)  & 14.56\scriptsize {(+9.28)}    & 72.59\scriptsize {(+18.21)}           & 16.23\scriptsize {(+3.53)}   & 73.29\scriptsize {(+21.65)}           & 21.65\scriptsize {(+13.59)}   & 81.33\scriptsize {(+16.82)}           & 12.94\scriptsize {(+4.63)}    & 83.11\scriptsize {(+12.06)}             \\
RGAT-BERT$_\texttt{IEGA}$ (100\%) & 16.49\scriptsize {(+11.21)}    & 72.80\scriptsize {(+18.42)}           & 16.25\scriptsize {(+3.55)}   & 75.65\scriptsize {(+24.01)}           & 23.50\scriptsize {(+15.44)}   & 82.95\scriptsize {(+18.44)}           & 13.60\scriptsize {(+5.29)}    & 85.09\scriptsize {(+14.04)}           \\
\hlineB{4}
\end{tabular}}
\vspace{-4mm}
\end{table*}

\vspace{-3mm}
\subsection{Experimental Analysis}
\label{sec:experimental analysis}
% First, we apply our \texttt{IEGA} framework to two classical ABSC model to explore the explanations of the models with few labeled opinion words. 
% Second, we compare these models with several ABSC models to evaluate the performance of our framework. 
% Third, we comfirm the robustness of our model on adversarial dataset.
\vspace{-1mm}
\textbf{Interpretability Analysis. } We apply our \texttt{IEGA} framework to two classical ABSC models to explore the explanations of the models with different proportions of labeled opinion words (Fig. \ref{fig:results of interpretability} and Table \ref{table:results of interpretability_2}). 
The results show that our model captures the explanations (opinion words) more accurately. For example, both BERT-SPC$_\texttt{IEGA}$ and RGAT-BERT$_\texttt{IEGA}$ obtain higher HR and MRR, which indicates that they find opinion words more effectively than the corresponding models without \texttt{IEGA}. 
This is because the model can better capture the opinion words corresponding to the aspect with the help of gradient correction. 
For Post-hoc Accuracy, we compute the accuracy by selecting the top $5$ words based on their importance weights to make a prediction.
%Our model gains an increase of 10 points over Lap14 and Res16 in terms of Posthoc accuracy with only 10\% (about 150) training samples.
Our model gains an increase of five points over all datasets in terms of Post-hoc Accuracy using only 10\% (about 150) training samples annotated with opinion words. 
Also, models with \texttt{IEGA} perform better than ones without \texttt{IEGA} in terms of AOPC despite only partial training samples labeled opinion words. 
% Thus, if deleting the most related words (opinion words) obtained respectively by the models with \texttt{IEGA} and the models without \texttt{IEGA}, the ones with \texttt{IEGA} reduces more performance while models without \texttt{IEGA} is still focusing on words that may not matter. 
In summary, the improvement of these metrics shows that the interpretability of the model can be largely boosted with the help of our \texttt{IEGA} framework even if only 10\% of the opinion words are annotated. 
%Also, for AOPC, we can find that deleting the most related words obtained by the models with \texttt{IEGA} reduces more performance than the ones without \texttt{IEGA}. The improvement of these metrics shows that the interpretability of the model can be largely boosted with the help of our \texttt{IEGA} framework. 
% AOPC calculates the average change of accuracy over test data by deleting top $K$ words via word-level saliency map.

\begin{table}[t!]
% \vspace{-5mm}
\centering
\small
\caption{Performance of our models and baselines.}
\label{table:results of acc}
\vspace{-2mm}
\setlength{\tabcolsep}{0.5mm}{\begin{tabular}{lcccccccc}
\hlineB{4}
\multirow{2}{*}{Model}   & \multicolumn{2}{c}{Lap14} & \multicolumn{2}{c}{Res14} & \multicolumn{2}{c}{Res15} & \multicolumn{2}{c}{Res16} \\
                         & Acc.    & F1    & Acc.    & F1    & Acc.    & F1    & Acc.    & F1    \\
                         \hline
AEN-BERT                 & 81.80       & 56.07       & 88.59       & 64.90       & 86.44       & 63.73       & 88.60       & 65.06       \\
LCF-BERT                 & 81.83       & 58.23       & 90.00       & 72.69       & 85.94       & 67.53       & 89.91       & 69.98        \\
BERT-SPC                 & 81.07       & 62.84       & 89.34       & 67.91       & 85.02       & 56.44       & 88.02       & 56.23       \\
RGAT-BERT                & 82.58       & 65.10       & 91.64       & 77.50       & 87.09       & 69.36       & 90.78       & 67.34       \\ \hline
BERT-SPC$_\texttt{IEGA}$  & 82.28       & 62.93       & 90.62       & 72.75       & 85.40       & 59.39       & 88.56       & 62.60       \\
\scriptsize{Improvement}  & \scriptsize{(+1.21)}       & \scriptsize{(+0.09)}       & \scriptsize{(+1.28)}       & \scriptsize{(+4.84)}       & \scriptsize{(+0.38)}       & \scriptsize{(+2.95)}       & \scriptsize{(+0.54)}       & \scriptsize{(+6.37)}       \\
RGAT-BERT$_\texttt{IEGA}$ & \textbf{83.08}       & \textbf{65.56}       & \textbf{92.36}       & \textbf{79.30}       & \textbf{88.25}       & \textbf{72.49}       & \textbf{91.76}       & \textbf{76.02}       \\
\scriptsize{Improvement}              & \scriptsize{(+0.50)}             & \scriptsize{(+0.46)}             & \scriptsize{(+0.72)}             & \scriptsize{(+1.80)}            & \scriptsize{(+1.17)}            & \scriptsize{(+3.13)}            & \scriptsize{(+0.98)}            & \scriptsize{(+8.68)}    \\
\hlineB{4}
\end{tabular}}
\vspace{-3mm}
\end{table}

\textbf{Performance Analysis. }
We compare our models with several ABSC models to evaluate the performance of our framework (Table \ref{table:results of acc}). From the results, we obtain the following observations. \textbf{First}, our model performs better than the baselines over all the datasets in terms of accuracy and F1. RGAT-BERT$_\texttt{IEGA}$ obtains the best results compared with all the existing state-of-the-art baselines (e.g., RGAT-BERT). \textbf{Second}, our \texttt{IEGA} framework can improve the performance of the base model. 
% Particularly, the model with \texttt{IEGA} performs better than the corresponding models without \texttt{IEGA}.
For instance, F1 improved by 3 and 8 points on Res15 and Res16 by integrating $\texttt{IEGA}$ with RGAT-BERT.
% \subsection{Interpretability}
% \label{sec:interpretability}

\begin{table}[t!]
% \vspace{-5mm}
\centering
\small
\caption{The results of robustness analysis.}
\label{table:results of robustness}
\vspace{-2mm}
\begin{tabular}{clcccccccc}
\hlineB{4}
& \multirow{2}{*}{Model}   & \multicolumn{2}{c}{AddDiff} & \multicolumn{2}{c}{RevNon}       \\
              &           & Acc.    & F1    & Acc.    & F1    \\
                         \hline
\multirow{2}{*}{Lap14} & BERT-SPC                 & 45.21       & 32.01       & 50.20       & 41.61\\
& BERT-SPC$_\texttt{IEGA}$          & 48.74       & 35.81       & 52.47       & 44.47 \\ \hline
\multirow{2}{*}{Res14} & BERT-SPC                 & 64.84       & 50.98       & 62.44       & 41.61\\
& BERT-SPC$_\texttt{IEGA}$ & 67.10       & 52.28       & 66.03       & 58.16\\ \hline
\multirow{2}{*}{Res15} & BERT-SPC                 & 46.38       & 33.38       & 56.58       & 38.45\\
& BERT-SPC$_\texttt{IEGA}$ & 53.26       & 39.22       & 58.00       & 40.94\\ \hline
\multirow{2}{*}{Res16} & BERT-SPC                 & 43.04       & 36.57       & 56.95       & 43.05\\
 & BERT-SPC$_\texttt{IEGA}$ & 52.19       & 39.68       & 59.07       & 43.60\\
\hlineB{4}
\end{tabular}
\vspace{-2mm}
\end{table}

\textbf{Robustness Analysis. }
We also analyze the robustness of our proposed \texttt{IEGA} framework (Table \ref{table:results of robustness}).
We test our model on two robustness testing datasets released by TextFlint \cite{wang2021textflint}. RevNon reverses the sentiment of the non-target aspects with originally the same sentiment as the target aspect's, and AddDiff adds a new non-target aspect whose sentiment is opposite to the target aspect of the sentence.
We find that BERT-SPC$_\texttt{IEGA}$ outperforms BERT-SPC over all datasets in terms of accuracy and F1. 
It shows that the model infers the sentiment based on the opinion words w.r.t the aspect stably. 
% Thus it will not be affected by perturbations in the training data. 
These observations suggest that our framework does have a large improvement in the robustness of the model.

\vspace{-4mm}
\section{Related Work}
\label{sec:related work}
\vspace{-2mm}
\textbf{Gradient-based Analysis Models. }
Recently, studies on explanation methods has grown, including perturbation-based \cite{ribeiro2016should}, gradient-based \cite{binder2016layer} and visualization-based \cite{zeiler2014visualizing} methods. 
We focus on the gradient-based method \cite{DBLP:journals/corr/SimonyanVZ13}, a popular algorithm applicable to different neural network models. 
Gradient-based methods \cite{DBLP:journals/corr/GoodfellowSS14}
have been widely applied into CV and NLP \cite{zeiler2014visualizing,DBLP:conf/ijcai/0002LSBLS18}. 
% There are three common saliency approaches: Simple Gradient \cite{DBLP:journals/corr/SimonyanVZ13}, Smooth Gradient \cite{DBLP:journals/corr/SmilkovTKVW17}, Integrated Gradients \cite{sundararajan2017axiomatic}.
The gradient-based approach is also used to understand the predictions of the text classification models from the token level \cite{li2016visualizing,alikaniotis2016automatic}. 
In addition, Rei et al. \cite{rei2018zero} adopted a gradient-based approach to detect the critical tokens in the sentence via the sentence-level label. 
The existing work also found that the gradient-based models are easily manipulable \cite{wang2020gradient} and unreliable \cite{kindermans2019reliability}.
In this paper, we design an \texttt{IEGA} algorithm to force the model to discover the target-aware opinion words using the gradient.

\textbf{Aspect-based Sentiment Classification. }
In recent years, thanks to the introduction of pre-trained language models, it has made tremendous progress in many NLP tasks, including aspect-based sentiment classification (ABSC) \cite{zhou2019deep}. By simply connecting the aspect words with the original sentence and then inputting them into BERT for training, researchers obtain excellent results in ABSC tasks \cite{xu-etal-2019-bert}.
Furthermore, Song et al.\cite{song2019attentional} proposed AEN-BERT, which adopts attention-based encoders to model the interaction between context and aspect. Zeng et al. \cite{zeng2019lcf} proposed LCF-BERT, which uses multi-head self-attention to make the model focus on the local context words. Wang et al. \cite{wang-etal-2020-relational} proposed a relation-aware graph attention network to model the tree structure knowledge for sentiment classification. 
However, most of them focus on improving performance, while the explanation of ABSC is not well studied.
Yadav et al. \cite{yadav2021human} proposed a human-interpretable learning approach for ABSC, but there is still a big gap in accuracy compared to the state-of-the-art methods.

% \vspace{-3mm}
\section{Conclusions and Future Work}
\vspace{-2mm}
\label{sec:conclusions and future work}
In this paper, we introduce an \texttt{IEGA} framework to improve the explanations of the gradient-based methods for ABSC. We design a gradient correction algorithm based on the word-level saliency map via a tiny amount of labeled samples. We conduct extensive experimental results with various metrics over four popular datasets to verify the interpretation, performance, and robustness of the models using our \texttt{IEGA}. 
We also explore the influence of sample numbers and find that our framework can effectively improve the interpretation with small samples.
It would be interesting to explore the performance of our model on more existing methods and tasks because \texttt{IEGA} is model agnostic and task agnostic.

\vspace{-2mm}
\section*{Acknowledge}
\vspace{-2mm}
The authors wish to thank the reviewers for their helpful comments and suggestions. 
This research is funded by the National Key Research and Development Program of China (No. 2021ZD0114002), the National Natural Science Foundation of China (No. 61907016) and the Science and Technology Commission of Shanghai Municipality Grant (No. 22511105901 \& No. 21511100402). 

\vfill\pagebreak
\newpage\clearpage
\small
\bibliographystyle{IEEEbib}
\bibliography{refs}

\end{document}